\documentclass[10pt,twocolumn,letterpaper]{article}

\usepackage[pagenumbers]{cvpr} 
\usepackage{float}
\definecolor{cvprblue}{rgb}{0.21,0.49,0.74}
\usepackage[pagebackref,breaklinks,colorlinks,allcolors=cvprblue]{hyperref}
\usepackage{tikz}
\def\paperID{16863} 
\def\confName{CVPR}
\def\confYear{2025}

\title{Uncertainty-Aware Regression for Socio-Economic Estimation via  Multi-View Remote Sensing}

\author{
  Fan Yang$^{1}$ \qquad Sahoko Ishida$^{1}$ \qquad Mengyan Zhang$^{1}$ \qquad Daniel Jenson$^{1}$ \\
  Swapnil Mishra$^{2}$ \qquad Jhonathan Navott$^{1}$ \qquad Seth Flaxman$^{1}$ \\[2mm]
  $^{1}$University of Oxford \qquad $^{2}$National University of Singapore \\
  {\tt\small \{fan.yang, sahoko.ishida, mengyan.zhang\}@cs.ox.ac.uk, daniel.jenson@worc.ox.ac.uk,} \\
  {\tt\small swapnilmishra.anu@gmail.com, jhonathan.navott@worc.ox.ac.uk, seth.flaxman@cs.ox.ac.uk}
}

\begin{document}
\maketitle
\begin{abstract}
Remote sensing imagery offers rich spectral data across extensive areas for Earth observation. Many attempts have been made to leverage these data with transfer learning to develop scalable alternatives for estimating socio-economic conditions, reducing reliance on expensive survey-collected data. However, much of this research has primarily focused on daytime satellite imagery due to the limitation that most pre-trained models are trained on 3-band RGB images. Consequently, modeling techniques for spectral bands beyond the visible spectrum have not been thoroughly investigated. Additionally, quantifying uncertainty in remote sensing regression has been less explored, yet it is essential for more informed targeting and iterative collection of ground truth survey data. In this paper, we introduce a novel framework that leverages generic foundational vision models to process remote sensing imagery using combinations of three spectral bands to exploit multi-spectral data. We also employ methods such as heteroscedastic regression and Bayesian modeling to generate uncertainty estimates for the predictions. Experimental results demonstrate that our method outperforms existing models that use RGB or multi-spectral models with unstructured band usage. Moreover, our framework helps identify uncertain predictions, guiding future ground truth data acquisition.
\end{abstract}

\begin{figure*}[t]
  \centering
  \includegraphics[width=0.8\textwidth]{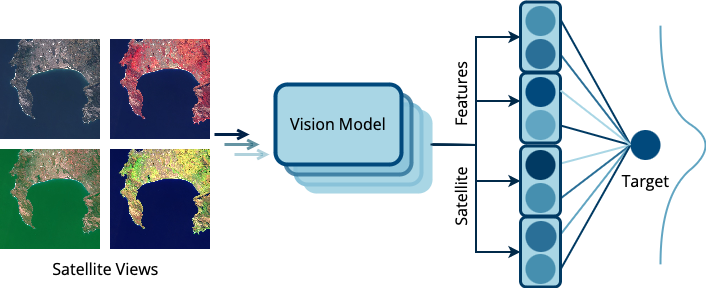}
  \caption{We select distinct \emph{satellite views} of the same location across different band groups, capturing unique spatial features. These views are then processed through separate pre-trained vision models to extract satellite features, which are subsequently aggregated to predict the target variable with associated uncertainty estimates.}
  \label{fig:main_fig}
\end{figure*}
\section{Introduction}
\label{sec:intro}

Satellite imagery is a crucial tool for detailed observation across a variety of studies, including environmental science \cite{burke2021sustain, chang2018env_mon}, agriculture \cite{nguyen2020satagri, yang2013agr, you2017crop}, and urban monitoring \cite{furberg2020saturban, Kussul2017satlandclassif}. Programs like Landsat, launched by NASA, and Copernicus, initiated by the European Space Agency (ESA), provide high-resolution satellite imagery that is publicly accessible for both research and industrial purposes. These initiatives offer time-stamped visual data of the Earth's surface, archiving regional snapshots over extended periods. The datasets are rich, containing multi-spectral information that extends beyond the visible spectrum, with comprehensive global coverage and frequent imaging intervals. Such features facilitate the use of advanced remote-sensing techniques, particularly deep learning, which supports the accurate extraction of geospatial features and recognition of patterns in the regions of interest.

The majority of remote sensing research in computer vision focuses on tasks such as classification, segmentation, object detection, cloud removal, and imagery generation. These tasks are fundamental to extracting meaningful information from satellite imagery and have been largely driven by convolutional neural network (CNN) architectures due to their exceptional performance in capturing spatial features \cite{cheng2020rsisc, guo2018objcnn}. Classification is widely used to categorize land cover types and land use \cite{Kussul2017satlandclassif, naushad2021landuse}, while segmentation techniques are applied to delineate different regions within an image \cite{tian2021segmentation}, such as urban areas, water bodies, or vegetation. Object detection aims to locate and identify specific items, like vehicles, buildings, or natural features, making it valuable for urban planning and environmental monitoring \cite{nurkarim2023objfootprint, adegun2023objmodels}. Cloud removal involves generative modeling essential to recover missing data and ensure usable imagery, as cloud cover often obstructs important features \cite{singh2018cloudgan, zhao2021cloudgan}.

Regression analysis using satellite imagery, on the other hand, is less common in remote sensing yet is useful for the estimation of various socioeconomic indicators, such as population density or economic activity. These analyses predominantly rely on daylight imagery and pre-trained CNN models to infer relationships between visual features and socioeconomic variables \cite{jean2016cnnpov, yeh2020sat_cnn_africa}. Despite their effectiveness, CNNs have several limitations in this context. They struggle to capture global patterns due to their reliance on fixed receptive fields. Moreover, CNNs require a fixed input size, which often requires down-sampling when dealing with high-resolution satellite images of varying dimensions. Recent developments in satellite imagery foundational models, such as SatMAE \cite{cong2022satmae} and SatMAE++ \cite{satmaepp2024rethinking}, leverage vision transformer backbones and support multi-spectral capabilities. However, the spectral bands are combined without clear intuition for the grouping structure, where the model could struggle to identify useful patterns.

Given the current state of research, our work provides the following methods targeting the estimation of socioeconomic indicators with remote sensing data sources. Beyond the use of daylight RGB imagery, our approach incorporates other bands in the multi-spectral data to create visually three-band combinations, which we refer to as ``views". These views are designed to emphasize different landscape features, such as vegetation, geology, and water bodies, providing diverse perspectives of the same region. By leveraging different spectral bands, we are able to capture a broader range of information, enhancing the analysis beyond what is possible with standard RGB imagery.

As illustrated in Figure \ref{fig:main_fig}, we utilize distinct pre-trained transformer-based vision models to process each view in parallel. This approach allows us to form a comprehensive satellite feature representation of the imagery, which can be used for subsequent tasks. The transformer-based model's ability to capture global context and integrate information across different views enables more robust feature extraction compared to traditional CNN-based models. We integrate different visual features from distinct views to form more comprehensive feature representations. We experimentally evaluate the model's performance on the KidSat \cite{sharma2024kidsat} dataset that links remote sensing with childhood poverty prediction, comparing it with the best-performing satellite regression model in the benchmark. To improve the applicability of our method and facilitate greater social impact, we also provide an uncertainty estimation pipeline regarding our modeling framework to offer guidance for data collectors and policymakers in designing targeted survey sampling strategies to obtain future ground truth data more efficiently.

Our contributions are summarized as follows:

\begin{itemize}
    \item We incorporate multiple spectral bands to form meaningful three-band views, highlighting different features such as vegetation, geology, and water bodies.
    \item We leverage pre-trained transformer-based vision models to process each view in parallel to generate comprehensive satellite feature representations for downstream regression.
    \item We provide uncertainty estimation of the modeling framework to provide additional insights into the model's prediction.
\end{itemize}

\section{Related Work}
\label{sec:related_work}

\paragraph{Regression with Remote Sensing.} The use of satellite imagery to infer continuous variables has become a novel method of geospatial modeling, with various approaches demonstrating its potential to assess socioeconomic, demographic, and environmental conditions, particularly in data-scarce regions. Early works utilized night-time light intensity as a proxy for economic activity, leveraging the correlation between brightness and economic output \cite{henderson2012nightlightecon, bennet2017ntl}. Efforts have been made to extend this approach by using pre-trained CNNs to learn from both daytime satellite images and night-time lights data, demonstrating the usage of visual imagery by predicting poverty in African countries\cite{jean2016cnnpov,xie2015transfer, yeh2020landsat}. More intuitively, the visual patterns in satellite imagery such as building and infrastructure can be recognized for efficient population estimation \cite{Tiecke2017building, robinson2017buildingpop}. Extending on this, researchers have also explored using remote sensing to infer population dynamics \cite{zong2019deepdpm} and assess exposure to climate hazards \cite{tellman2021satfloodpop}.

\paragraph{Representation Learning in Remote Sensing.} Representation learning has shown significant potential in the context of satellite imagery, where models are developed to convert raw images into useful feature representations. These feature representations encapsulate essential information from the original images, simplifying subsequent tasks such as regression analysis and enabling more efficient downstream modeling. One prominent approach for statistical representation of satellite data is MOSAIKS (Multi-task Observation using SAtellite Imagery \& Kitchen Sinks) \cite{rolf2021mosaiks}, which encodes satellite imagery into generalizable feature representations, allowing various downstream applications with a shared, reusable feature space. Deep learning has further expanded the possibilities of representation learning for satellite imagery. Generic vision models, such as MAE (Masked Autoencoder) \cite{he2022mae} and DINO (Distillation with NO Labels) \cite{caron2021dino}, pre-trained on standard RGB data, can be adapted to extract meaningful features from visible spectra of satellite images. Moreover, specialized satellite versions of these models have been developed to address unique challenges in remote sensing. SatMAE \cite{cong2022satmae}, SatCLIP \cite{klemmer2023satclip}, and DINO-MC \cite{wanyan2024dinomc} are examples of representation learning models extended to satellite imagery. Notably, SatMAE utilizes a masked autoencoder architecture and is capable of handling multiple spectral bands as grouped channels. However, deploying the pre-trained SatMAE requires resizing satellite imagery to the pre-trained configuration (224 by 224 pixels), resulting in a potential compromise in imagery quality or a reduction in geospatial coverage when working with high-resolution data. Recent research \cite{sharma2024kidsat} has also explored the application of these models to predicting child poverty. Comparisons between DINOv2 \cite{oquab2024dinov2} and SatMAE, both using single imagery without time labels, demonstrate the utility of these representation learning approaches in extracting socioeconomic insights from satellite data.

\paragraph{Combination of Imagery Bands.}

One key advantage of remote sensing data is its collection across multiple wavelengths, including visible, near-infrared, and short-wave infrared spectra. The use of 3-band combinations in Sentinel imagery visualizations has been explored to offer different perspectives of the same location, highlighting distinct properties such as natural color views, crop growth, and soil condition \cite{lemenkova2020sentinelcomb}. Several recent studies have demonstrated that beyond the standard RGB bands, coastal blue, near-infrared (NIR) and short-wave infrared (SWIR) bands prove useful in tasks like land cover classification \cite{kulkarni2021bandlandcover}, permafrost landform mapping \cite{bhuiyan2020bandlandpermafrost}, and agriculture uses \cite{zhang2017bandagri}.

\begin{figure}[t]
    \centering
    \begin{subfigure}[t]{0.48\linewidth}
        \centering
        \includegraphics[width=1\textwidth]{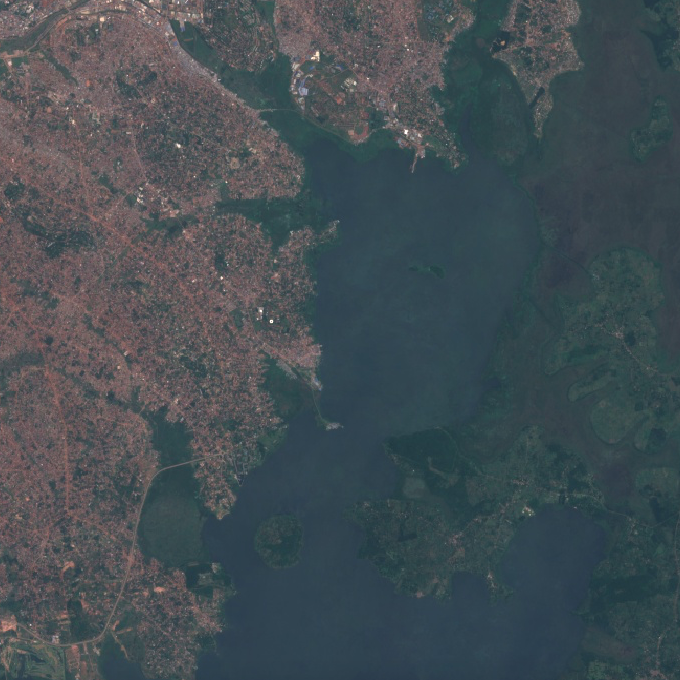}
        \caption{Natural  Color View.}
        \label{fig:sub1}
    \end{subfigure}
    \hfill
    \begin{subfigure}[t]{0.48\linewidth}
        \centering
        \includegraphics[width=1\textwidth]{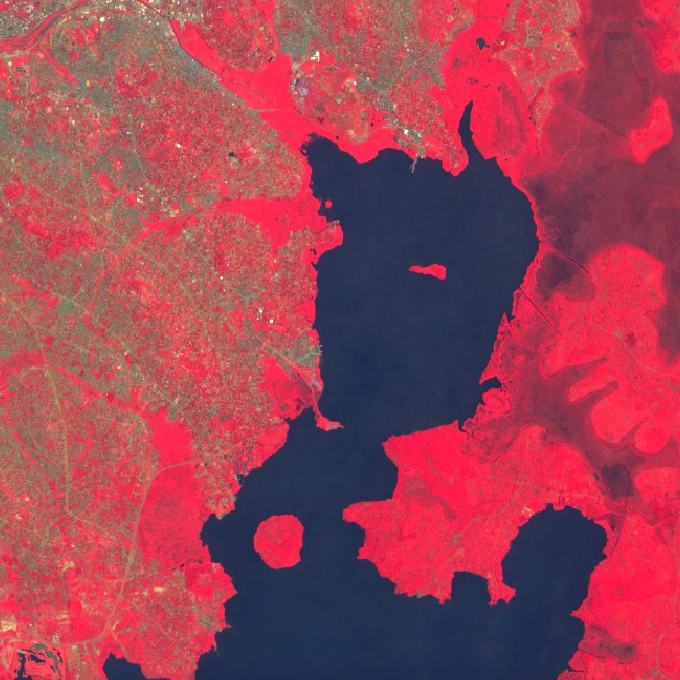}
        \caption{Fault Color View.}
        \label{fig:sub2}
    \end{subfigure}
    
    \vskip\baselineskip
    
    \begin{subfigure}[b]{0.48\linewidth}
        \centering
        \includegraphics[width=1\textwidth]{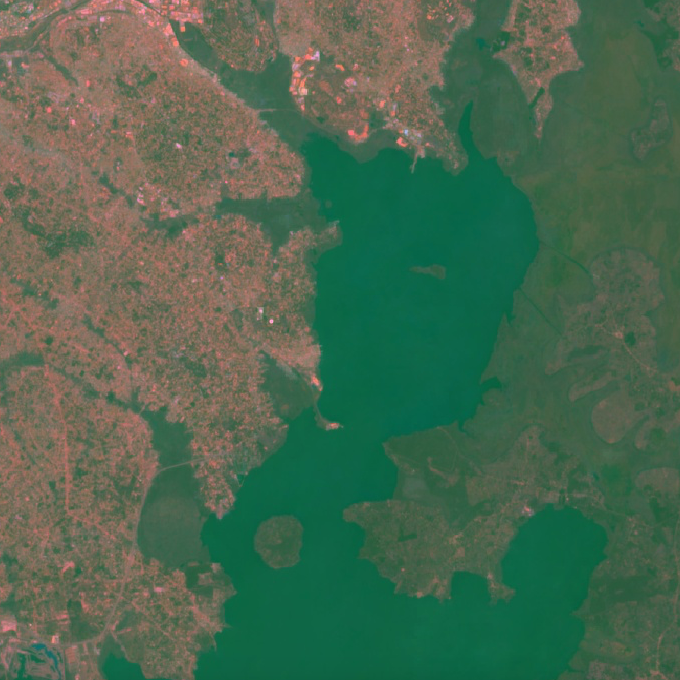}
        \caption{Land Moisture View.}
        \label{fig:sub3}
    \end{subfigure}
    \hfill
    \begin{subfigure}[b]{0.48\linewidth}
        \centering
        \includegraphics[width=1\textwidth]{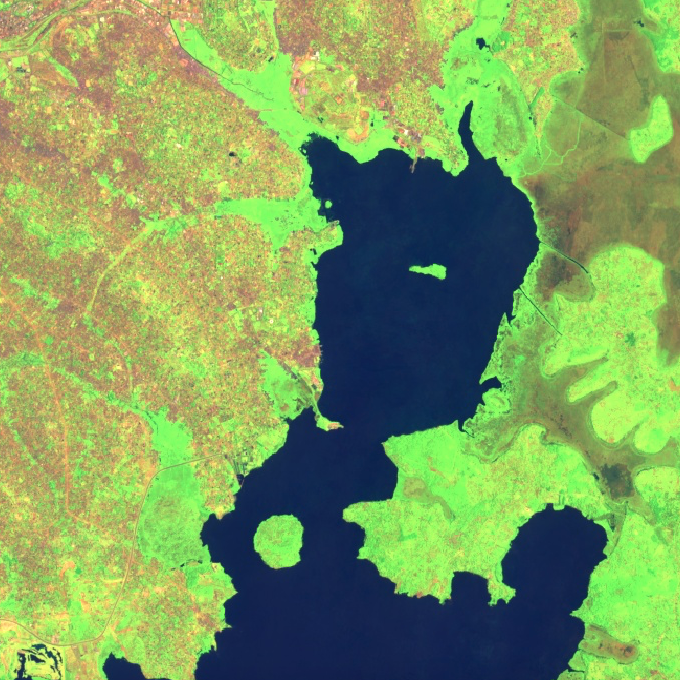} 
        \caption{Agriculture View.}
        \label{fig:sub4}
    \end{subfigure}
    
    \caption{Examples of different views near Murchison Bay, Uganda. The natural color view uses bands B4 (Red), B3 (Green), and B2 (Blue). The false-color view incorporates bands B8 (Near Infrared), B4 (Red), and B2 (Blue). The land moisture view is composed of bands B12 (Short-Wave Infrared), B1 (Coastal Blue), and B3 (Green). Lastly, the agriculture view includes bands B11 (Short-Wave Infrared), B8 (Near Infrared), and B2 (Red).}
    \label{fig:views}
\end{figure}

Figure \ref{fig:views} illustrates how different combinations of spectral bands can highlight distinct aerial features. The natural color view represents the standard visible spectrum, providing general visual information about urban infrastructure, vegetation cover, and water bodies. The false-color view emphasizes vegetation health, using the NIR band (B8) to reflect chlorophyll levels, which distinguishes conditions of vegetation health. The land moisture view shows moisture content through various shades of green. The agriculture view highlights vegetation health, where bright green indicates vigorous growth, and non-crop features, such as mature trees, appear as muted green. Figure \ref{fig:views} particularly demonstrates the case that in the natural color view, the water body and vegetation are less distinguishable, while the false color and agriculture views provide better contrast to highlight distinct landscapes. This illustrates the importance of multi-view analysis in extracting satellite features from each location.

\section{Method}

\subsection{Multi-view Modeling}

Our proposed method extracts features from Sentinel-2 satellite imagery by leveraging spectral diversity to enhance the representation of landscape characteristics. Unlike the benchmark methods approach by \citet{sharma2024kidsat}, which utilizes RGB imagery with Vision Transformers (ViTs), our approach employs a multi-view imagery processing scheme that integrates information from distinct perspectives, aiming to capture a broader range of features across landscapes. 

Formally, let $ X_{i,v} \in \mathbb{R}^{H \times W \times 3} $ represent the 3-band composite imagery data for a location $i$ and view $v \in \{1,\dots,V\}$, where $H$ and $W$ are the height and width of the image. Each view corresponds to a different combination of spectral bands, such as natural color, false color, moisture content, and agricultural analysis.

\paragraph{Feature Extraction Using Vision Transformers}
Each 3-band composite is processed through a Vision Transformer (ViT) block to extract feature representations for each view. Let $ Z_{i,v} = \text{ViT}_{v}(X_{i,v}) \in \mathbb{R}^{1 \times d} $, where $Z_{i,v}$ represents the aggregated representation of all patches, and $d$ is the dimensionality of the features extracted from each patch. Each view $v$ is handled using a Vision Transformer block $ \text{ViT}_{v} $, which learns to focus on different aspects of the landscape.
\paragraph{Fine-tuning}

The fine-tuning process of satellite imagery representation $Z_{i,v}$ utilizes the target variable or a set of auxiliary indicators and covariates from survey data that are associated with the target variable. These auxiliary indicators provide additional supervisory signals that help fine-tune $Z_{i,v}$ to reduce overfitting.

We pass $Z_{i,v}$ through a linear transformation followed to predict the target/auxiliary indicators:

\begin{equation*}
    \hat{\mathbf{s}}_i = Z_{i,v} \mathbf{W}_v + \mathbf{b}
\end{equation*}

where $\mathbf{W}$ and $\mathbf{b}$ are the weights and biases of the linear transformation.

The model is trained to minimize the difference between the predicted target/auxiliary indicators $\hat{\mathbf{s}}_i$ and the ground truth indicators $\mathbf{s}_i$ using the L1 or L2 Loss function:

\begin{equation*}
    \mathcal{L} = \frac{1}{N} \sum_{i=1}^N \left\lVert \hat{\mathbf{s}}_i - \mathbf{s}_i \right\rVert
\end{equation*}

where $N$ is the total number of locations, and $\left\lVert \cdot \right\rVert$ denotes the distance from the prediction to the true label.

By minimizing $\mathcal{L}$, we fine-tune the parameters of the model, to yield the final representation $Z'_{i,v}$, to better predict the target/auxiliary indicators.

\paragraph{Aggregation for Multi-view Fusion}
To generate a comprehensive representation of each location, we concatenate the feature representations from all views. Formally, we define:

\[
\mathbf {F}_i = Z'_{i,1} \oplus Z'_{i,2} \oplus \dots \oplus Z'_{i,V} \in \mathbb{R}^{P \times Vd}
\]

where \(\mathbf {F}_i \) is the final combined representation obtained by concatenating these updated feature representations. This approach effectively integrates complementary information from all \( V \) views to capture the complexity of the landscape.

\paragraph{Prediction of the Target Variable}

If the fine-tuning process is not directly optimizing the prediction on the target variable, a ridge regression with cross-validation (RidgeCV) for hyperparameter tuning is then used to predict the target variable from the encoder part of the model. Ridge regression effectively handles multicollinearity in high-dimensional data such as imagery features generated by large vision models. The prediction is given by:

\begin{equation}
    \hat{y}_i = \mathbf{F}_i \mathbf{w} + b
\label{linear}
\end{equation}

where $\mathbf{w}$ and $b$ are the regression weights and bias, respectively, and $\hat{y}_i$ is the predicted value of the target variable at location $i$.

\subsection{Uncertainty Estimation}

The proposed framework, which encodes imagery into feature representations using deep learning models, allows for various methods for uncertainty quantification of the predictions. We focus on two primary techniques for estimating model uncertainty: heteroscedastic uncertainty estimation using deep learning and Bayesian linear regression.

\paragraph{Heteroscedastic Uncertainty Estimation Using Deep Learning}

Heteroscedastic uncertainty estimation leverages deep learning models to predict both the mean and variance of the target variable as functions of the input features. For each input imagery \( x \), the model outputs a mean \( \mu(x) \) and a variance \( \sigma^2(x) \):

\[
y \sim \mathcal{N}(\mu(x), \sigma^2(x))
\]

Assuming that the observation noise is Gaussian and its variance depends on the input (heteroscedastic), the likelihood of observing \( y \) given \( x \) is:

\[
p(y|x) = \frac{1}{\sqrt{2 \pi \sigma^2(x)}} \exp \left( -\frac{(y - \mu(x))^2}{2 \sigma^2(x)} \right)
\]

The negative log-likelihood (NLL) loss function for the dataset \( \mathcal{D} = \{(x_i, y_i)\}_{i=1}^N \) is then:

\begin{align}
\label{equ: NLL}
    \text{NLL} = \sum_{i=1}^N \left( \frac{(y_i - \mu(x_i))^2}{2 \sigma^2(x_i)} + \frac{1}{2} \log \sigma^2(x_i) \right)
\end{align}

By minimizing \( \text{NLL} \), the model learns to adjust both \( \mu(x) \) and \( \sigma^2(x) \) to fit the data to capture input-dependent noises. This is particularly useful in scenarios where the uncertainty varies across different regions of the input space.

For a new input \( x^* \), the predictive distribution is:

\[
y^* \sim \mathcal{N}(\mu(x^*), \sigma^2(x^*))
\]

where \( y^* \) is the predicted value for \( x^* \). The mean and variance of the prediction are:

\[
\mathbb{E}[y^*] = \mu(x^*)
\]
\[
\text{Var}[y^*] = \sigma^2(x^*)
\]

This method allows the model to provide both point estimates and uncertainty measures that reflect the heteroscedastic nature of the data; however, this method does not capture epistemic uncertainty, which would require the use of Bayesian deep learning models to account for.

\paragraph{Bayesian Linear Regression}
Bayesian linear regression provides a probabilistic framework for modeling the relationship between encoded satellite features and the target variable. With the linear model defined in \ref{linear}, assuming Gaussian noise \(\varepsilon_i \sim \mathcal{N}(0, \sigma^2)\), the observed target \(y_i\) is:
\[
y_i = \tilde{\mathbf{F}}_i \tilde{\mathbf{w}} + \varepsilon_i
\]
where \(\tilde{\mathbf{F}}_i = [\mathbf{F}_i,\, 1]\) and \(\tilde{\mathbf{w}} = \begin{bmatrix} \mathbf{w} \\ b \end{bmatrix}\). 
With a Gaussian prior on \(\tilde{\mathbf{w}}\):
\[
\tilde{\mathbf{w}} \sim \mathcal{N}(\mathbf{0}, \Omega_{\tilde{\mathbf{w}}})
\]
and the likelihood of the data:
\[
p(y_i | \tilde{\mathbf{w}}, \tilde{\mathbf{F}}_i) = \mathcal{N}(y_i | \tilde{\mathbf{F}}_i \tilde{\mathbf{w}}, \sigma^2),
\]
the posterior distribution of \(\tilde{\mathbf{w}}\) given the data \(\mathcal{D}\) is Gaussian. A common choice for the covariance matrix of the prior is \({\Omega}_{\tilde{\mathbf{w}}}=c\mathbf{I}\), with \(c>0\), which corresponds to Bayesian ridge regression. However, when the number of features is large, it may be more appropriate to use a prior that induces greater sparsity. This will be discussed in \ref{section: experiment uncertainty}.

The predictive distribution for a new input \(\tilde{\mathbf{F}}^*\) is:
\[
p(y^* | \tilde{\mathbf{F}}^*, \mathcal{D}) = \mathcal{N}\left( \tilde{\mathbf{F}}^{*\top} \boldsymbol{\mu}_{\tilde{\mathbf{w}}},\, \tilde{\mathbf{F}}^{*\top} \boldsymbol{\Sigma}_{\tilde{\mathbf{w}}} \tilde{\mathbf{F}}^* + \sigma^2 \right)
\]
where \(\boldsymbol{\mu}_{\tilde{\mathbf{w}}}\) and \(\boldsymbol{\Sigma}_{\tilde{\mathbf{w}}}\) are the posterior mean and covariance of \(\tilde{\mathbf{w}}\). 


\section{Experiments}

In this section, we first introduce the dataset and benchmark used to evaluate our multi-view modeling approach. We then provide detailed explanations of our implementation, specifically for the fine-tuning and evaluation stages applied to the selected dataset. Notably, we demonstrate the advantages of transfer learning compared to training from scratch. We also compare our model’s performance to existing best-performing models on the dataset, utilizing the full spectrum of available spectral bands. Finally, we present our results, including an evaluation of the quality of uncertainty estimation in our predictions.

\subsection{Evaluation Dataset}

Due to the limited dataset and benchmark available for socio-economic variables and high-resolution remote sensing, we focused the evaluation of our methods on the KidSat \cite{sharma2024kidsat} work. The KidSat dataset is a challenging benchmark for evaluating satellite-based feature representations for multidimensional child poverty estimation. It includes 33,608 satellite images from 16 countries in Eastern and Southern Africa, covering 1997 to 2022. These images are paired with high-quality survey data on child poverty from Demographic and Health Surveys (DHS). Unlike typical tasks like land cover classification, estimating child poverty from satellite imagery presents unique challenges due to the abstract nature of poverty, which encompasses factors like housing, water, sanitation, nutrition, education, and health.

\paragraph{Satellite Imagery}
The KidSat dataset includes both Landsat and Sentinel-2 imagery paired with DHS survey variables. We select Sentinel-2 imagery as the input source for vision models to generate feature representations due to its higher spatial resolution (up to 10 m/pixel) and additional red-edge bands, which emphasize vegetation health and chlorophyll content. For each survey location, the input data is a 994 x 994 pixel imagery with 13 spectral bands, covering a 10 km x 10 km region. A detailed description of the spectral bands and spatial coverage of Sentinel-2 imagery is provided in the Appendix. 

\paragraph{Evaluation Target}The dataset utilizes the ``severe deprivation" variable as the target variable for assessing model performance. This variable quantifies the proportion of children experiencing severe deprivation within a given survey area, expressed as a normalized value ranging between 0 and 1. Model efficacy is measured through the use of mean absolute errors (MAEs), which calculate the percentage discrepancy between the model predictions and the actual data.

\paragraph{Fine-tuning Scheme}The formulation of the severe deprivation variable involves extracting and aggregating 17 distinct sub-indicators from the DHS survey data. These sub-indicators are then normalized and transformed into a 99-dimensional vector that represents various aspects of childhood poverty. Detailed methodologies for processing these indicators can be found in the Appendix. In this study, we experiment with two training/fine-tuning schemes: the first scheme directly optimizes the prediction of the ``severe deprivation" variable, while the second approach refines the model using the 99-dimensional poverty vector, subsequently applying this refined model to predict the severe deprivation outcome.

\subsection{End-to-End Training vs. Fine-Tuning}

\begin{figure}[t]
    \centering
    \includegraphics[width=\linewidth]{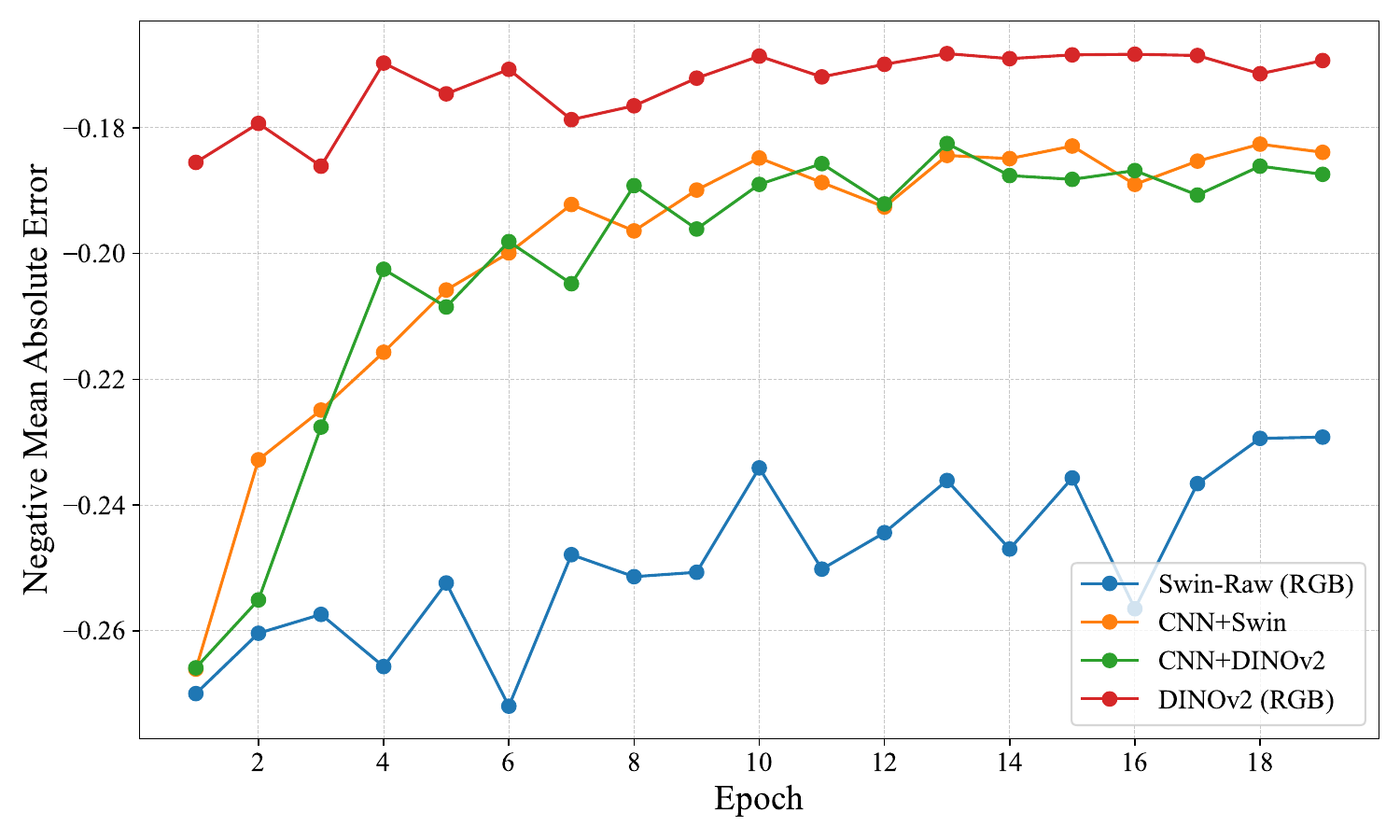}
    
    \caption{Validation errors for predicting severe deprivation using various model configurations: a raw SWIN model accepting RGB imagery, a pre-trained SWIN model and a pre-trained DINOv2-ViT-Base model with a CNN head that maps 13-channel imagery to 3 channels, and a pre-trained DINOv2-ViT-Base model accepting RGB imagery. Performance is tracked by negative mean absolute error over the fine-tuning process.}
    \label{fig:e2e_vs_ft}
\end{figure}

 In this section, we investigate the impact of adapting transfer learning to the end-to-end training of vision models. The training process follows a general uni-view scheme: imagery with dimensions $B * H * W$, where $B$ represents the number of bands, and $H$ and $W$ represent the height and width, respectively, is fed into the vision model to extract satellite features. A linear transformation is then applied to these features to map them to the target variable, severe deprivation, using a sigmoid activation function to model the percentage nature of severe deprivation. We use a 80-20 split on the dataset to generate the training and validation sets and record the validation error using mean absolute error (MAE) along the training process.
\subsubsection{Configuration}
In the transfer learning experiment, we utilize the ViT-Base model pre-trained using the DINOv2 framework \cite{oquab2024dinov2} and the pre-trained SWIN transformer \cite{liu2021Swin}. One of the DINOv2-ViT-Base models is provided with RGB imagery, and the other is provided with full Sentinel-2 imagery by appending a convolutional layer to transform multi-spectral imagery to 3-band imagery. For the SWIN models, one is initialized as a raw model to accept RGB imagery, and the other uses a CNN head that accepts 13-band imagery and is loaded with pre-trained weights. The learning rate and weight decay are set to 1e-6 for fine-tuning DINOv2, while a learning rate of 1e-5 is used for fine-tuning SWIN transformers. All models are trained using the Adam optimizer, with L1 loss and a batch size of 2, on an A100 GPU for 20 epochs.
 
\subsubsection{Discussion} 
Figure \ref{fig:e2e_vs_ft} presents the performance on the validation split measured by mean absolute error. Our results show a significant improvement when using a pre-trained model compared to training from scratch. Specifically, the DINOv2 model—pre-trained on the ImageNet and designed for RGB representation learning—quickly converges to a low error, outperforming the raw SWIN transformer model. We observe that applying a simple CNN transformation to map multi-spectral imagery to a pre-trained model improves upon the raw SWIN model. However, this approach involves automatic mixing of the multi-spectral data, which does not structurally exploit the rich spectral information available. Consequently, it is less competitive than fully leveraging a pre-trained model with limited spectral bands, as demonstrated by the DINOv2-RGB results.


\subsection{Multi-view Fine-tuning}

We experimented with the proposed multi-view fine-tuning framework by selecting four views presented in Figure \ref{fig:views}: (i) B4 (Red), B3 (Green), B2 (Blue); (ii) B8 (Near Infrared), B4 (Red), B2 (Blue); (iii) B12 (Short-Wave Infrared), B1 (Coastal Blue), B3 (Green); and (iv) B11 (Short-Wave Infrared), B8 (Near Infrared), B2 (Red). The satellite features encoded from these four views are to fine-tuned with the 99-dimensional poverty vector. The prediction involves mapping the concatenated vectors to predict the target variable using ridge regression. Other models for comparison are also fine-tuned on the poverty vector. The performance is reported using mean absolute errors.

\subsubsection{Configuration}

We selected the pre-trained DINOv2 ViT-Base model as our imagery encoder for the multi-view experiment. The learning rate and weight decay are both set to 1e-6. For comparison, we used DINOv2 with ViT-Base and ResNet-50 backbone and RGB-only imagery under the same learning rate and weight decay settings. SatMAE with its pre-trained model accepts 224×224 imagery, while other models are fed with 994×994 imagery. We used 1e-5 as the learning rate for SatMAE. We used the Adam optimizer, with L1 loss and a batch size of 2, and the experiments were carried out on A100 GPUs.

\begin{table}
  \centering
  \begin{tabular}{@{}lc@{}}
    \toprule
    Model & MAE $\pm$ SE \\
    \midrule
    Mean Predictor & 0.2930 $\pm$ 0.0031  \\
    SatMAE-GC & 0.1993 $\pm $ 0.0015 \\
    DINO-ResNet50 & 0.2399 $\pm$ 0.0009\\
    DINOv2-ViT-Base & 0.1663 $\pm$ 0.0023\\
    DINOv2-ViT-Base Multi-view & \textbf{0.1605 $\pm$ 0.0020}\\
    \bottomrule
  \end{tabular}
  \caption{Experimental results of estimating severe deprivation variable, results reported in mean absolute error and standard error is calculated using 5-fold cross-validation. SatMAE-GC uses 10 spectral bands grouped as \cite{cong2022satmae}, DINOv2-ResNet50 and DINOv2-ViT-Base receive RGB imagery, and DINO-ViT-Base Multi-view receives four 3-band views.}
    \label{tab:maes}
    
\end{table}

\subsubsection{Discussion}

We present our results in Table \ref{tab:maes}. The baseline is established by predicting the mean value of the training dataset, providing a benchmark for absolute error. The pre-trained model with a CNN-based backbone achieves a Mean Absolute Error (MAE) of 0.2399, which surpasses the baseline but underperforms compared to models based on ViT. SatMAE, pre-trained on 224x224 satellite imagery using multi-spectral data, achieves an MAE of 0.1993. DINOv2-ViT, utilizing RGB imagery at full resolution, further improves upon this result with an MAE of 0.1663. By employing a multi-view fine-tuning scheme, the model reduces the prediction error even further to 0.1605.

Table \ref{tab:view-perform-cv} presents the performance of individual views across cross-validation folds. We observe that while the model with RGB imagery (B4, B3, B2) generally stands out with the most accurate prediction in most folds, other views can be more informative in certain data mixes (e.g., fold 3). However, the multi-view model, which leverages concatenated feature vectors from all views, consistently outperforms each individual view in every cross-validation fold.

This finding aligns with the intuition illustrated in Figure \ref{fig:views}, where different views of the same location present distinct landscape appearances. The satellite imagery dataset contains locations with diverse landscape features, and a single view may fail to capture the full extent of these characteristics. By aggregating features from multiple views, the multi-view approach provides a more comprehensive representation of the data, while the downstream ridge regression helps to regularize the increased dimensionality of the feature vector.

\begin{table}
  
  \centering
  \begin{tabular}{lcccccc}
    \toprule
    View & Fold 1 & Fold 2 & Fold 3 & Fold 4 & Fold 5\\
    \midrule
    4,3,2 & \textbf{0.1587} & \textbf{0.1645} & 0.1719 & \textbf{0.1726} & \textbf{0.1638}  \\
    8,4,2 & 0.1651 & 0.1679 & \textbf{0.1649} & 0.1781 & 0.1838  \\
    12,1,3 & 0.1611 & 0.1705 & 0.1702 & 0.1789 & 0.1800\\
    11,8,2 & 0.1643 & 0.1737 & 0.1749 & 0.1801 & 0.1810  \\
    \midrule
    All & 0.1554 & 0.1606 & 0.1620 & 0.1680 & 0.1564 \\
    \bottomrule
  \end{tabular}
  \caption{Comparison of MAE for severe deprivation from an individual perspective across each fold of the cross-validation. Combining feature representations from multiple views consistently outperforms predictions based on any single view, while no individual view’s features consistently outperform others across all cross-validation folds.}
  \vspace{1mm}
  \label{tab:view-perform-cv}
\end{table}

\subsection{Uncertainty Estimation} \label{section: experiment uncertainty}

\begin{figure*}[t]
    \centering
    \begin{subfigure}[t]{0.3\textwidth}
        \centering
        \includegraphics[width=1\textwidth]{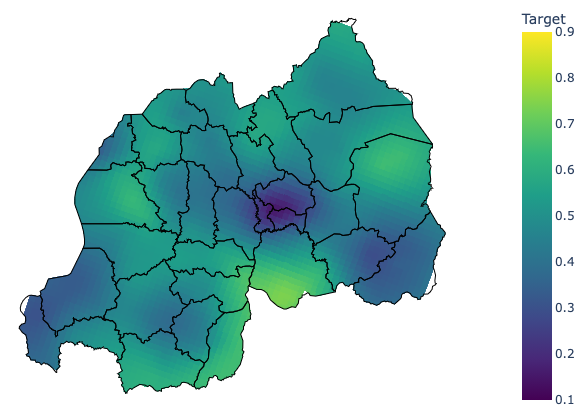}
        \caption{Severe Deprivation Variable.}
        \label{fig:target}
    \end{subfigure}
    \begin{subfigure}[t]{0.3\textwidth}
        \centering
        \includegraphics[width=1\textwidth]{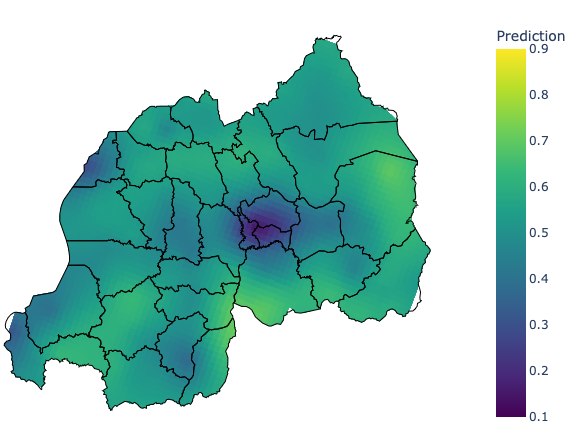}
        \caption{Multi-view Prediction.}
        \label{fig:predict}
    \end{subfigure}
    \begin{subfigure}[t]{0.3\textwidth}
        \centering
        \includegraphics[width=1\textwidth]{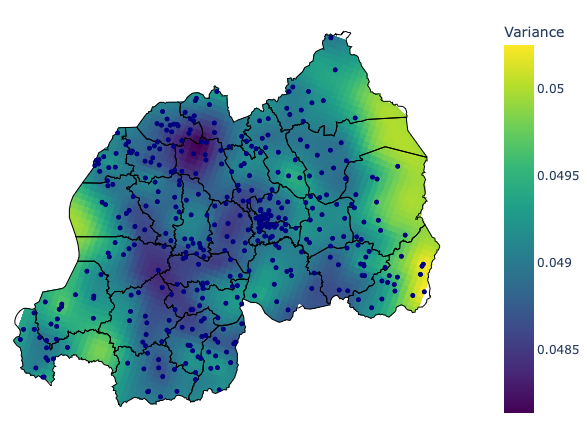}
        \caption{Uncertainty Estimation Using BLR.}
        \label{fig:uncer}
    \end{subfigure}
    
    \caption{Example Predictions for Rwanda in 2019. Figure (a): Ground truth labels derived from the DHS dataset, visualized using Ordinary Kriging. Figure (b): Posterior mean estimates obtained using DINOv2-ViT-Base with a multi-view fine-tuning scheme and Bayesian linear regression. Figure (c): Posterior variance from the Bayesian linear regression, with blue markers indicating locations where training data exist.}
    \label{fig:heatmap}
\end{figure*}

The estimation of prediction uncertainty is performed using heteroscedastic uncertainty estimation and Bayesian linear regression. The experiment setup is conducted and demonstrated on the multi-view DINOv2-ViT-Base model.

\subsubsection{Configuration}
\paragraph{Heteroscedastic Regression}
We adopt a simple approach where, instead of only predicting the target variable, we predict its mean and variance. These predictions are fed into a negative log-likelihood loss function, allowing the model to backpropagate and update the mean and variance. The model is fine-tuned using the same parameter setup as previously described on NVIDIA GH200 superchips.

\paragraph{Bayesian Linear Regression}
To manage the high dimensionality of features, we apply a shrinkage prior to the regression coefficients in the regression model \ref{linear}. The prior on \({\mathbf{w}}=(w_1,\ldots,w_{Vd})\) is specified as:
\[
w_j \sim \mathcal{N}(0,\lambda_j^2\tau^2), 
\]
\[
\lambda_j\sim t_\nu^+(0,1)
\]
where \(t_\nu^+\) denotes half-student-\(t\) prior with degrees of freedom \(\nu\). Setting \(\nu=1\) results in the horseshoe prior \cite{carvalho2009handling}, while increasing \(\nu\) reduces sparsity but mitigates a common issue with horseshoe prior: divergent transitions during sampling. We selected \(\nu=3\) and compared this with an alternative shrinkage prior, the regularized horseshoe prior \cite{piironen2017sparsity}, observing only minor differences in predictive performance and uncertainty evaluation. Our approach is fully Bayesian, incorporating a half-Cauchy prior on \(\tau\). We use a non-sparsifying, weakly informative prior for the intercept term,\(b\sim \mathcal{N}(0,5^2)\), and a flat prior on \(\sigma\). The model was implemented in the probabilistic programming language Stan \cite{Stan} and sampled using four chains, each with 1,500 draws, discarding the initial 500 as warm-up draws.

\subsubsection{Uncertainty Evaluation Metrics}\label{section: uncertainty metrics}

We employ the following metrics to evaluate uncertainty:

\begin{itemize}
\item Interval Length and Coverage: For a specified quantile level  $0 < \alpha < 1$ , we assess the percentage of true values captured within the  $(100 \times \alpha)\%$  prediction intervals of the posterior distribution, alongside the corresponding interval length. Higher coverage indicates better reliability in capturing true values, while shorter interval lengths reflect more precise uncertainty estimates. Ideally, we aim for a balance where coverage is sufficiently high and interval length is minimized.

\item Negative Log-Likelihood (NLL): The NLL (defined in Eq. \ref{equ: NLL}) is a commonly used metric for quantifying the quality of uncertainty estimates. It consists of two components: a penalty for prediction errors scaled by the model's confidence and a penalty for overconfidence when the model underestimates uncertainty. Lower NLL values indicate better overall performance, reflecting accurate predictions with well-calibrated uncertainty.

\item Continuous Ranked Probability Score (CRPS): CRPS measures the squared distance between the predictive distribution and the empirical distribution associated with the observed value. The lower the CRPS, the better the alignment between the predicted distribution and the actual observations.
\begin{align*}
    \text{CRPS}\left(P, x_a\right) = \int_{-\infty}^{\infty} \left\| P(x) - H\left(x - x_a\right) \right\|_2 \, dx,
\end{align*}
where $x_a$ is the true observed value, $P(x)$ denotes the predicted cumulative distribution function, and $H(x)$ is the Heaviside step function.

\end{itemize}

\subsubsection{Result}
Table \ref{tab:uncertainty evaluation} shows the results of the uncertainty evaluation for the two methods across the specified metrics. Bayesian Linear Regression (BLR) with a shrinkage prior outperforms Heteroscedastic Regression (HR) on all metrics, achieving a shorter interval length, matching the theoretical 95\% coverage, and yielding lower values for both NLL and CRPS. However, the interval length for BLR remains relatively wide, indicating room for improvement in precision. Additionally, with only two methods compared, further investigation with additional models is warranted for a more comprehensive assessment of uncertainty estimation.

\begin{table}[]
    \centering

\begin{tabular}{lcccc}
\hline
    & Interval length & Coverage & NLL    & CRPS  \\ \hline
HR & 0.955           & 0.750    & 1.309  & 0.247 \\
BLR & 0.712           & 0.951    & -0.101 & 0.123 \\ \hline
\end{tabular}  
    \caption{Comparison of uncertainty estimation methods using Bayesian Linear Regression (BLR) and Heteroscedastic Regression (HR). Metrics include the length and coverage of the 95\% predictive interval, Negative Log-Likelihood (NLL), and Continuous Ranked Probability Score (CRPS).}
    \label{tab:uncertainty evaluation}
\end{table}

\section{Conclusion}

Satellite imagery is a special form of image data that often consists of spectral bands beyond the standard RGB channels. Existing state-of-the-art models are not structurally engineered to integrate and process the relationships between these additional spectral bands. Multi-spectral satellite imagery provides valuable insights into an area’s characteristics, such as vegetation content and moisture levels, which can serve as implicit indicators for socio-economic variables, while current methods do not intentionally model these indicators. Our work bridges recent advances in traditional remote sensing, which typically uses three-band views, with foundational vision deep learning models to extract information from multiple spectral perspectives. We demonstrate our approach to the estimation of childhood poverty and show that our model outperforms existing methods that use only RGB imagery or group multiple spectral bands without intuitive structure. Future directions for this work include pre-training foundational models on satellite imagery with structured combinations of spectral bands to enhance their ability to capture landscape features.
\newpage

{
    \small
    \bibliographystyle{ieeenat_fullname}
    \bibliography{main}
}

\clearpage
\setcounter{page}{1}
\setcounter{section}{0} 
\renewcommand{\thesection}{\Alph{section}} 
\maketitlesupplementary

\section{Code}
\label{sec:code}
The code used to implement and reproduce the experiments described in this work is available on GitHub. It includes scripts for indicator preprocessing, model training, and evaluation, as well as instructions for replicating the experimental setup. The repository can be accessed through this \href{https://github.com/lukeyf/multiview_remote_sensing}{GitHub repository}.

\section{Satellite Imagery}

In this work, we examine the model's performance on childhood poverty prediction using \textbf{Sentinel-2} imagery. Sentinel-2 Imagery is a mission by the European Space Agency (ESA) under the Copernicus Programme, designed to provide high-resolution optical imagery for land monitoring. The mission comprises two identical satellites, Sentinel-2A and Sentinel-2B, launched in June 2015 and March 2017, respectively. Operating in the same orbit but phased at 180° to each other, they offer a revisit time of five days at the equator, enhancing the temporal resolution for continuous observation.

The satellites capture imagery across 13 spectral bands, ranging from visible light to shortwave infrared (SWIR), as detailed in Table \ref{sentinel}. This multispectral capability, combined with high spatial resolution, makes Sentinel-2 particularly valuable for applications in agriculture, forestry, land cover classification, and disaster management.

\begin{table}[hbtp]
    \centering
    \begin{tabular}{@{}lccc@{}}
    
        \toprule
        \textbf{Band Name} & \textbf{Index} & \textbf{Wavelength} & \textbf{Resolution} \\
        \midrule
        Coastal Aerosol & 1  & 443  & 60 \\
        Blue            & 2  & 494  & 10 \\
        Green           & 3  & 560  & 10 \\
        Red             & 4  & 665  & 10 \\
        Red Edge 1      & 5  & 703  & 20 \\
        Red Edge 2      & 6  & 740  & 20 \\
        Red Edge 3      & 7  & 782  & 20 \\
        NIR & 8  & 835  & 10 \\
        NIR Narrow      & 8A & 864  & 20 \\
        Water Vapour    & 9  & 945  & 60 \\
        SWIR 1          & 11 & 1610 & 20 \\
        SWIR 2          & 12 & 2190 & 20 \\
        Cirrus          & 10 & 1375 & 60 \\
        \bottomrule
    \end{tabular}
    \caption{This table lists the Sentinel-2 spectral bands with their central wavelengths (in nanometers) and spatial resolutions (in meters per pixel). NIR denotes Near-Infrared bands, and SWIR denotes Short-Wave Infrared bands.}
    
    \label{sentinel}
\end{table}

\subsection{Key Characteristics of Sentinel-2 Imagery}

\begin{itemize}
\item \textbf{Spatial Resolution}: Varies by band—10 meters for visible (Blue, Green, Red) and NIR bands; 20 meters for Vegetation Red Edge and SWIR bands; and 60 meters for atmospheric correction bands (Coastal Aerosol, Water Vapour, SWIR—Cirrus).
\item \textbf{Temporal Resolution}: A revisit time of five days at the equator with both satellites operational, enabling frequent and consistent monitoring.
\item \textbf{Spectral Range}: 13 spectral bands covering the visible, near-infrared, and shortwave infrared regions, providing comprehensive spectral information for diverse applications.
\item \textbf{Swath Width}: A wide swath of 290 kilometers, allowing extensive area coverage in a single pass.
\item \textbf{Radiometric Resolution}: 12-bit data depth, offering pixel values ranging from 0 to 4095, which enhances the detection of subtle reflectance differences.
\end{itemize}

\subsection{Considerations for Data Processing}

The high spatial and spectral resolution of Sentinel-2 imagery offers detailed and rich information crucial for large-scale vision models and precise analytical tasks. However, this richness comes with increased computational demands. Processing such high-resolution data can be resource-intensive and time-consuming, especially when dealing with large geographic extents or extensive temporal datasets. Efficient data processing strategies, including data reduction techniques and optimized algorithms, are essential to balance the trade-off between data quality and computational feasibility.

Moreover, Sentinel-2’s operational timeframe, starting from June 2015, means it does not provide imagery for events or changes that occurred before this date. This limitation should be considered when conducting historical analyses or long-term environmental monitoring.

We adopt the imagery normalization approach used in Google Earth Engine to preprocess the satellite data. For Sentinel-2 imagery, we scale the original pixel values from a range of 0 to 3000 down to 0 to 255, clipping any values outside this initial range. This scaling preserves the relative intensity of the pixel values while adapting them for image rendering.

\end{document}